\documentclass{article}

\usepackage[utf8]{inputenc} % allow utf-8 input
\usepackage[T1]{fontenc}    % use 8-bit T1 fonts
\usepackage{hyperref}       % hyperlinks
\usepackage{url}            % simple URL typesetting
\usepackage{booktabs}       % professional-quality tables
\usepackage{amsfonts}       % blackboard math symbols
\usepackage{nicefrac}       % compact symbols for 1/2, etc.
\usepackage{microtype}      % microtypography
\usepackage{lipsum}
\usepackage{fancyhdr}       % header
\usepackage{graphicx}       % Required for inserting images
\usepackage[square, numbers, sort]{natbib}
\graphicspath{{media/}}     % organize your images and other figures under media/ folder

\title{Human Understanding AI Paper Challenge 2024 - Dataset Design}

\author{
Se Won Oh, Hyuntae Jeong, Jeong Mook Lim,\\ Seungeun Chung, Kyoung Ju Noh\\
\texttt{\{sewonoh, htjeong, jmlim21, schung, kjnoh\}@etri.re.kr}\\
Electronics and Telecommunications Research Institute\\
Daejeon, South Korea\\
}

\date{March 2024}

\begin{document}
\maketitle

\begin{abstract}
In 2024, we will hold a research paper competition (the third Human Understanding AI Paper Challenge) for the research and development of artificial intelligence technologies to understand human daily life. This document introduces the datasets that will be provided to participants in the competition, and summarizes the issues to consider in data processing and learning model development. 
\end{abstract}

\section{Introduction}
In order to comprehend the multifaceted nature of human behavior in everyday contexts, it is essential to systematically collect and analyze extensive lifelogs obtained from multimodal sensors and user records. During the year 2020, we accumulated in excess of 10,000 hours of sensor data from a cohort consisting of 22 participants \cite{etrilifelog2020, chung2022real}, employing both smartphones and smartwatches for data acquisition. Further, we organized two paper competitions, held in 2022 \cite{challenge_first} and 2023 \cite{challenge_second} respectively, with the primary objective of stimulating diverse studies aimed at leveraging artificial intelligence methodologies for the analysis of the aforementioned dataset.

For this year's competition, we are focusing on studies exploring the impact of various daily experiences on an individual's sleep patterns, emotional states, and stress levels. We also collected longer-term lifelog data in 2023 using a similar experiment protocol to 2020, and plan to make it available for this competition. In essence, the objective is to analyze sensor data gathered from smartphones and smartwatches to develop a robust learning model capable of discerning and inferring a comprehensive set of seven indicators linked to sleep quality, emotional responses, and stress levels.

\section{Dataset composition}
\label{chap:dataset_composition}
The data provided for the 2024 paper competition comprises three distinct datasets, as summarized in Table~\ref{tab:dataset_organization}. The train dataset incorporates information gathered in 2020 \cite{etrilifelog2020}, while the validation and the test datasets encompass additional data collected separately in 2023. Consequently, the demographic makeup of the participants associated with the train dataset differs from that of the participants associated with the validation and the test datasets. On the other hand, the cohort involved in constructing both the validation and the test datasets remains consistent. However, distinctive random integer identifiers(i.e. 'subject\_id') are assigned to each participant across the two datasets. Specifically, participant identifiers in the validation dataset range from 1 to 4, while those in the test dataset range from 5 to 8. Table~\ref{tab:demographics_2023} presents the demographics of each participant in the validation dataset. Each participant's age is presented in decade increments, while information regarding height and weight is rounded. 

\begin{table}[htb!]
    \caption{Dataset composition}
    \centering
    \begin{tabular}{c|c|c|c}
    \toprule
    Dataset & Number of participants & Days & Year of collection\\ 
    \midrule
    Train    & 22 & 508 & 2020\\ 
    Validation  & 4  & 105 & 2023\\ 
    Test     & 4  & 115 & 2023\\ 
    \bottomrule
    \end{tabular}
    \label{tab:dataset_organization}
\end{table}

\begin{table}[htb!]
    \caption{Demographics of the validation dataset}
    \centering
    \begin{tabular}{c|c|c|c|c|c}
    \toprule
    subject\_id & Gender & Age & Occupation & Height (cm)& Weight (kg)\\
    \midrule
    1 & male   & 50s & Self-employed & 168 & 88\\
    2 & female & 40s & Salaried & 163 & 62\\
    3 & male   & 20s & Unemployed & 178 & 74\\
    4 & female & 50s & Salaried & 155 & 59\\
    \bottomrule
    \end{tabular}
    \label{tab:demographics_2023}
\end{table}

\begin{figure}[htb!]
  \centering
  \includegraphics[trim=0 630 0 30, clip, width=0.8\columnwidth]{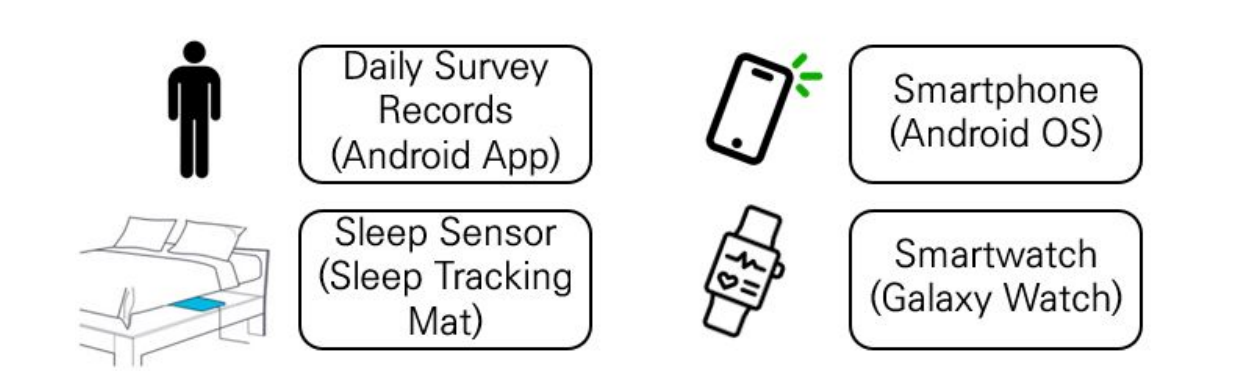}
  \caption{The sensors employed for lifelog collection}
  \label{fig:fig_sensors}
\end{figure}

Each dataset comprises the following four data items, as depicted in Figure~\ref{fig:fig_sensors}:
\begin{itemize}
\item Daily survey records: Self-recorded surveys completed either upon awakening or right before sleep each day.
\item Smartphone: Sensor data obtained through an application installed on a participant-owned smartphone (Android OS version 10 or higher).
\item Smartwatch: In the train dataset, acceleration and heart rate data were collected using an Empatica E4 device. For the validation and the test dataset, either a Galaxy Watch4 or Watch5 was employed to gather heart rate, step counts, and light intensity data.
\item Sleep sensor: Sleep log data from a Withings Sleep Tracking Mat\footnote{\url{https://www.withings.com/kr/en/sleep}}.     
\end{itemize}

\begin{table}[htb!]
    \caption{Key sensor data items and their corresponding data collection intervals}
    \centering
    \begin{tabular}{p{1.9cm}|p{4.1cm}|p{4.4cm}}
    \toprule    
    Data items  & \multicolumn{1}{c|}{Train Dataset} & \multicolumn{1}{c}{Validation and Test Dataset} \\
    
    \midrule
    Smartphone & 3-Axis acceleration (30Hz),\newline 
    Activity category (1/60Hz),\newline 
    GPS coordinates (1/5Hz),\newline 
    Application usage statistics\newline (every 30 minutes) & 3-Axis acceleration (\textbf{50}Hz),\newline 
    Activity category (1/60Hz),\newline 
    GPS coordinates (1/5Hz),\newline 
    Application usage statistics\newline (every \textbf{10} minutes),\newline
    \textbf{Light intensity} (1/600Hz),\newline
    \textbf{Ambient sound labels}\newline (1/120Hz)\\ \hline
    
    Smartwatch & 3-Axis acceleration (32Hz),\newline 
    Heart rate (1Hz) & \textbf{Step counts} (1/60Hz),\newline
    Heart rate (\textbf{1/60}Hz),\newline 
    \textbf{Light intensity} (1/600Hz)\\ \hline
    
    Daily survey\newline records & \raggedright{Subjective sleep quality\newline (5-point scale),\newline Emotion states (5-point scale),\newline Stress levels (5-point scale)} & Subjective sleep quality\newline (5-point scale),\newline Emotion states (11-point scale), \newline Stress levels (11-point scale)\\ \hline
    
    Sleep sensor & \multicolumn{2}{c}{Daily sleep log data (sleep duration, sleep efficiency, etc.)}\\
    \bottomrule
    \end{tabular}
    \label{tab:key_sensor_data_items}
\end{table}

Table~\ref{tab:key_sensor_data_items} provides an overview of the principal sensor data items included in the datasets, along with their respective time intervals for data collection. It is notable that the validation and the test datasets include additional data items that were not present in the train dataset. Additionally, it is crucial to acknowledge that the validation and the test dataset may contain a considerable amount of missing data. Comprehensive descriptions of the data configuration and format for the train dataset are available on the corresponding website and within the reference documentation\cite{etrilifelog2020, chung2022real}. Table~\ref{tab:detailed_sensor_data_items_in_2023} describes the detailed data configuration for the validation and the test datasets, emphasizing the data items collected from smartphones and smartwatches.

\begin{table}  [htb!]
    \centering 
    \caption{The detailed composition of the validation and the test datasets}
    \begin{minipage}{\linewidth}
    \begin{tabular}{c|p{2.5cm}|p{1.5cm}|p{4.3cm}}
    \toprule
    Name & \multicolumn{1}{c|}{Column} & \multicolumn{1}{c|}{DataType} & \multicolumn{1}{c}{Note}\\ 
    \midrule
    mAcc        & x\newline y\newline z\newline & float\newline float\newline float\newline & 3-Axis acceleration($m/{s}^{2}$)\\ \hline
    mActivity   & activity & string & Detected activity\footnote{0 (IN\_VEHICLE), 1 (ON\_BICYCLE), 2 (ON\_FOOT), 3 (STILL), 4 (UNKNOWN), 5 (TILTING), 7 (WALKING), 8 (RUNNING)}, classified by the Google ActivityRecognitionApi\footnote{\url{https://developers.google.com/location-context/activity-recognition}}, represented as an integer between 0 and 8\\ \hline
    mAmbience   & ambience\_labels & string & List of pairs of the top 10 labels and their probabilities (derived from AudioSet Ontology\footnote{\url{https://research.google.com/audioset/}})\\ \hline
    mGps        & altitude\newline latitude\newline longitude\newline speed & float\newline float\newline float\newline float\newline & Missing and unmeasured data are denoted by -1. The unit of speed is m/s.\\ \hline
    mLight      & m\_light & float & Ambient light in lx unit\\ \hline
    mUsageStats & m\_usage\_stats & string & List of app names and their respective usage times (in milliseconds unit)\\ \hline
    wHr         & heart\_rate & integer & Heart rate in beats per minute\\ \hline
    wPedo       & burned\_calories\newline distance\newline running\_steps\newline speed\newline steps\newline step\_frequency\newline walking\_steps\newline & float\newline float\newline integer\newline float\newline integer\newline float\newline integer\newline & Number of calories\newline Distance in meters\newline Number of running steps\newline Speed in m/s unit\newline Number of all steps\newline Step frequency in a minute\newline Number of walking steps\newline\\ \hline
    wLight      & w\_light & float & Ambient light in lx unit\\ 
    \bottomrule
    \end{tabular}
    \label{tab:detailed_sensor_data_items_in_2023}
    \end{minipage}    
\end{table}

Then we set up a series of classification problem wherein data collected from smartphones and smartwatches serve as the independent variables (X) for training a machine learning model, while metrics derived from either daily survey records or sleep sensor constitute the dependent variables (Y). 

\begin{figure}[htb!]
  \centering
  \includegraphics[trim=0 360 0 30, clip, width=0.5\linewidth]{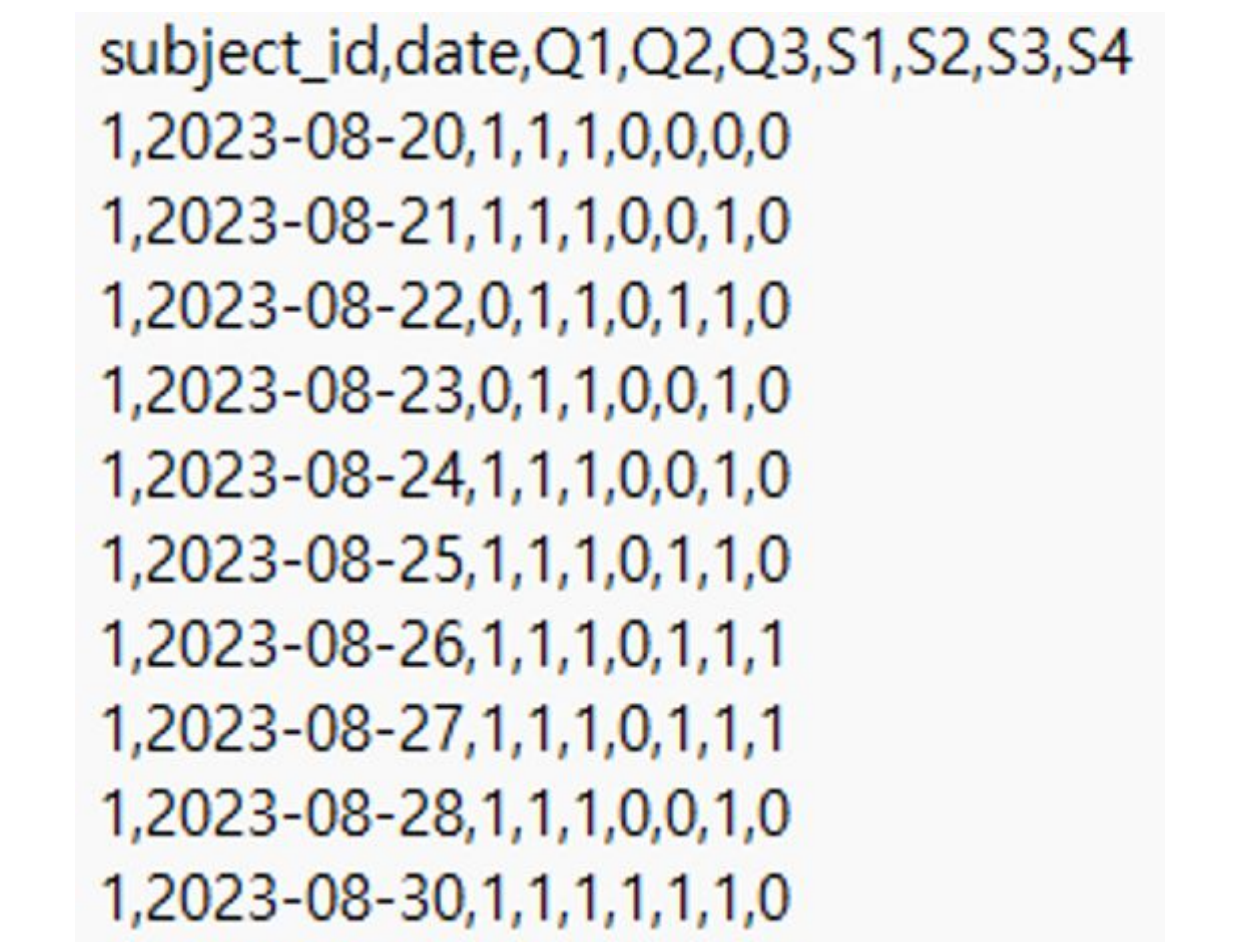}
  \caption{Example of the seven metrics}
  \label{fig:fig_metrics_sample}
\end{figure}

From the aforementioned data items of Table~\ref{tab:key_sensor_data_items}, we derive a total of seven metrics (Q1, Q2, Q3, S1, S2, S3, S4) based on the daily survey records and sleep sensor data, as described in Table~\ref{tab:Metrics}. Figure~\ref{fig:fig_metrics_sample} shows an example displaying the seven metrics per participant, per day, derived from the validation dataset.

In particular, the three questionnaire metrics (Q1, Q2, and Q3) related to the daily survey records are computed based on the average value of the questionnaire responses over the entire experimental period for each participant. For the first two questionnaire metrics (Q1 and Q2), a value of 1 is assigned to days with questionnaire responses above the average, while days with responses below the average receive a value of 0. The stress status indicator (Q3), conversely, assigns a value of 0 to days when each participant's stress level exceeds the average and a value of 1 to days when the stress level falls below the average.

\begin{table} [tb!]
    \centering
    \caption{Metrics to be identified}
    \begin{tabular}{c|p{6cm}|p{2.8cm}}
    \toprule 
    Metric & \multicolumn{1}{c|}{Explanation} & \multicolumn{1}{c}{Values}\\
    \midrule 
    Q1 & Overall sleep quality as perceived by a participant immediately after waking up & 0: below average,\newline 1: above average \\ \hline
    Q2 & Emotional state of a participant just before sleep & 0: below average,\newline 1: above average\\ \hline
    Q3 & Stress levels experienced by a participant just before sleep & 0: \textbf{above} average,\newline 1: \textbf{below} average\\ \hline
    S1 & Total sleep time (TST) & 0: below average,\newline 1: above average\\ \hline
    S2 & Sleep efficiency (SE) & 0: below average,\newline 1: above average\\ \hline
    S3 & Sleep onset latency (SOL, SL) & 0: below average,\newline 1: above average\\ \hline
    S4 & Wake after sleep onset (WASO) & 0: below average,\newline 1: above average\\ 
    \bottomrule
    \end{tabular}
    \label{tab:Metrics}
\end{table}

On the other hand, the four sleep metrics (S1, S2, S3, S4) derived from the sleep sensor data are calibrated to comply with the sleep health guidelines set forth by the National Sleep Foundation (NSF)\cite{nsf_guidelines}. These metrics assign a value of 1 to sleep records that adhere to the recommended guidelines, while sleep records not meeting the guidelines are assigned a value of 0.

All datasets described so far will be made available for registered participants in this competition to access and analyze.

\section{Submission format and evaluation method}

\begin{figure}[htb!]
  \centering
  \includegraphics[trim=0 30 0 30, clip, width=0.6\linewidth]{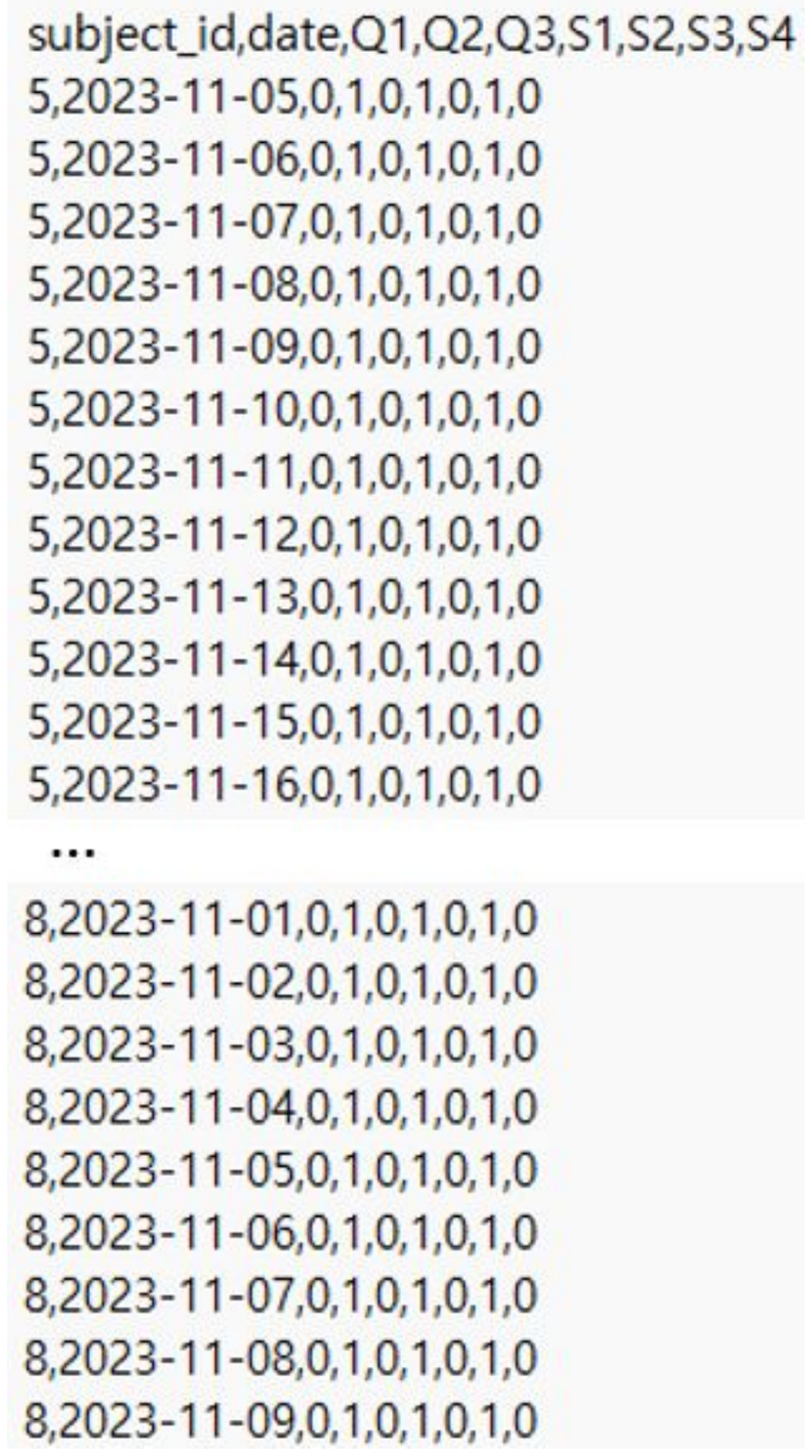}
  \caption{Example of a final submission file}
  \label{fig:fig_submission_example}
\end{figure}

The inference results of the learning model should be submitted in a comma-separated value (CSV) file format, including the participant IDs, dates, and the binary values (i.e., 0 or 1) for the seven metrics (Q1, Q2, Q3, S1, S2, S3, S4). As described in \autoref{chap:dataset_composition}, each participant's identifier must be represented as an integer value from 5 to 8, and the total data consists of 115 rows. Figure~\ref{fig:fig_submission_example} presents an example CSV submission file featuring participant IDs (subject\_id), dates, and seven metrics. 

Figure~\ref{fig:fig_submission_example} also implies that each metric necessitates binary classification, and the resulting assessment involves computing a macro F1 score for each metric followed by averaging these scores. However, it is important to consult the pertinent announcements regarding this year's competition, as the weighting of each of the seven metrics in the submitted work may vary, influencing the final ranking. Here, the F1 score serves as a means to consolidate the precision and recall of the model, calculated as the harmonic mean of the model's precision and recall values, as represented in equation (1).

\begin{equation}
F_{1}-score = 2\times \frac{Precision\times Recall}{Precision + Recall}
\end{equation}

\section{Conclusion}

This document introduces and details the designated datasets required for participants in the third Human Understanding AI Paper Challenge to analyze and develop learning models. We anticipate that this year's challenge would yield the development of best-performing learning models and techniques tailored to recognize various indicators of a person's daily life (i.e., sleep quality, emotional responses, and stress levels).

\section*{Acknowledgments}
This work was supported by Electronics and Telecommunications Research Institute (ETRI) grant funded by
the Korean government. [24ZB1100, Core Technology Research for Self-Improving Integrated Artificial Intelligence System].

%Bibliography
\bibliographystyle{unsrtnat}  
\bibliography{draft}  

\end{document}